\documentclass[runningheads]{llncs}
\usepackage{dsfont}
\usepackage{graphicx}
\usepackage{xcolor}
\usepackage{mathtools}
\usepackage{amsmath}
\DeclareMathOperator*{\argmax}{argmax}
\usepackage{amssymb}

\usepackage{tikz}

\usepackage[capitalize]{cleveref}
\crefname{section}{Sec.}{Secs.}
\Crefname{section}{Section}{Sections}
\Crefname{table}{Table}{Tables}
\crefname{table}{Tab.}{Tabs.}
\newcommand*\samethanks[1][\value{footnote}]{\footnotemark[#1]}

\begin{document}

\title{Bayesian Pseudo Labels: Expectation Maximization for Robust and Efficient Semi-Supervised Segmentation}

\titlerunning{Bayesian Pseudo Labels: EM for Semi Supervised Segmentation}
%
\author{
Mou-Cheng Xu\inst{1} \and
Yukun Zhou\inst{1} \and
Chen Jin\inst{1}\and
Marius de Groot\inst{2} \and
Daniel C. Alexander\inst{1} \and
Neil P. Oxtoby\thanks{Joint Senior Authorships.} \inst{1} \and
Yipeng Hu \samethanks \inst{1} \and
Joseph Jacob \samethanks \inst{1}
}

%
\authorrunning{Xu, Zhou, Jin, de Groot, Alexander, Oxtoby, Hu, Jacob}
%

\institute{
Centre for Medical Image Computing, University College London, UK \and
GSK R \& D, Stevenage, UK \and
\email{xumoucheng28@gmail.com, code: https://github.com/moucheng2017/EMSSL/}
}

\maketitle              
\begin{abstract}

This paper concerns pseudo labelling in segmentation. Our contribution is fourfold. Firstly, we present a new formulation of pseudo-labelling as an Expectation-Maximization (EM) algorithm for clear statistical interpretation. Secondly, we propose a semi-supervised medical image segmentation method purely based on the original pseudo labelling, namely SegPL. We demonstrate SegPL is a competitive approach against state-of-the-art consistency regularisation based methods on semi-supervised segmentation on a 2D multi-class MRI brain tumour segmentation task and a 3D binary CT lung vessel segmentation task. The simplicity of SegPL allows less computational cost comparing to prior methods. Thirdly, we demonstrate that the effectiveness of SegPL may originate from its robustness against out-of-distribution noises and adversarial attacks. Lastly, under the EM framework, we introduce a probabilistic generalisation of SegPL via variational inference, which learns a dynamic threshold for pseudo labelling during the training. We show that SegPL with variational inference can perform uncertainty estimation on par with the gold-standard method Deep Ensemble.

\end{abstract}

\section{Introduction}
Image segmentation is a fundamental component of medical image analysis, essential for subsequent clinical tasks such as computer-aided-diagnosis and disease progression modelling. In contrast to natural images, the acquisition of pixel-wise labels for medical images requires input from clinical specialists making such labels expensive to obtain. Semi-supervised learning is a promising approach to address label scarcity by leveraging information contained within the additional unlabelled datasets. Consistency regularisation methods are the current state-of-the-art strategy for semi-supervised learning.



Consistency regularisation methods \cite{meanteacher,fixmatch2020,remixmatch,mixmatch,Xu_MIDL} make the networks invariant to perturbations at the input level, the feature level \cite{cct_cvpr2020,bmvc_ssl_2020} or the network architectural level \cite{cpl_cvpr_2021}. These methods heavily rely on specifically designed perturbations at the input, feature or the network level, which lacks transferability across different tasks \cite{defense_pseudo_label}. In contrast, pseudo-labelling is a simple and general approach which was proposed for semi-supervised image classification. Pseudo-labelling \cite{PseudoLabel} generates pseudo-labels using a fixed threshold and uses pseudo-labels to supervise the training of unlabelled images. A recent work \cite{defense_pseudo_label} argued that pseudo labelling on its own can achieve results comparable with consistency regularisation methods in image classification. 

Recent works in semi-supervised segmentation with consistency regularisation are complex in their model design and training. For example, a recent work \cite{cct_cvpr2020} proposed a model with multiple decoders, where the features in each decoder are also augmented differently, before the consistency regularisation is applied on predictions of different decoders. Another recent method \cite{cpl_cvpr_2021} proposed an ensemble approach and applied the consistency regularisation between the predictions of different models. Other approaches include applying different data augmentation on the input in order to create two different predictions. One of the predictions is then used as a pseudo-label to supervise the other prediction \cite{fixmatch2020}. The aforementioned related works are all tested as baselines in our experiments presented in Sect.~\ref{sec:exps}.

\begin{figure}[t]
\centering
\includegraphics[width=0.8\textwidth]{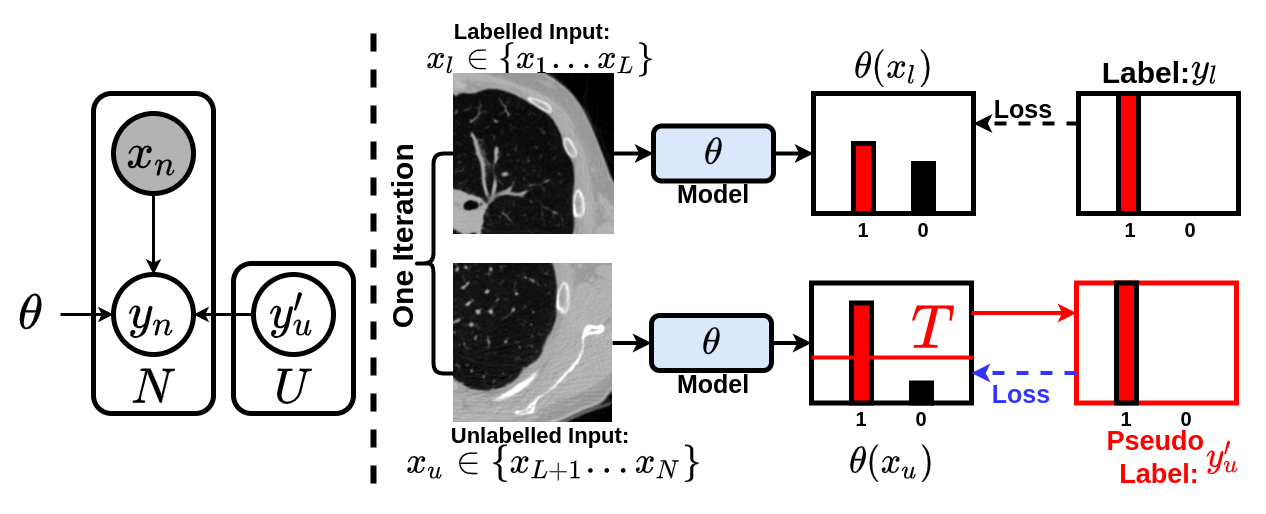}
\caption{Left: Graphical model of pseudo-labelling. Pseudo-label $y'_n$ is generated using data $x_u$ and model $\theta$, therefore, pseudo-labelling can be seen as the E-step in Expecation-Maximization. The M-step updates $\theta$ using $y'_n$ and data $X$. Right: Pseudo-labelling for binary segmentation. In our 1st implementation, namely SegPL, the threshold $T$ is fixed for selecting the pseudo labels. In our 2nd implementation, namely SegPL-VI, the threshold $T$ is dynamic and learnt via variational inference.}
\label{fig:main_method}
\end{figure}

In this paper, we propose a straightforward pseudo label-based algorithm called SegPL, and report its improved or comparable efficacy on two medical image segmentation applications following comparison with all tested alternatives. We highlight the simplicity of the proposed method, compared to existing approaches, which allows efficient model development and training. We also explore the benefits of pseudo labels at improving model generalisation with respect to out-of-distribution noise and model robustness against adversarial attacks in segmentation. Theoretically, we as well provide a new perspective of pseudo-labelling as an EM algorithm. Lastly, under the EM framework, we show that SegPL with variational inference can learn the threshold for pseudo-labelling leading to uncertainty quantification as well as Deep Ensemble \cite{deep_ensemble}. A high level introduction of SegPL can be found in Fig.\ref{fig:main_method}.

\section{Pseudo Labelling As Expectation-Maximization}
We show that training with pseudo labels is an Expecation-Maximization (EM) algorithm using a binary segmentation case. Multi-class segmentation can be turned into a combination of multiple binary segmentation via one-hot encoding and applying Sigmoid on each class individually (multi-channel Sigmoid).

Let $X= \{(x_1, y_1),... (x_L, y_L), x_{L+1},... x_{N} \}$ be the available training images, where $X_L = \{(x_l, y_l): l \in (1, ..., L)\}$ is a batch of $L$ labelled images; $X_U = \{x_u: u \in (L+1,..., N)\}$ is an additional batch of unlabelled images. Our goal is to optimise a segmentation network ($\theta$) so that we maximise the marginal likelihood of $P(X, \theta)$ w.r.t $\theta$. Given that pseudo-labelling is a generative task, we can cast pseudo-labels ($Y'_U=\{y'_u \in (L+1,..N)\}$) as latent variables. Now our optimisation goal is to maximise the likelihood of $P(X,Y'_U,\theta)$. We can solve $\theta$ and $Y'_U$ iteratively like the Expecation-Maximization (EM)\cite{bishop_ml} algorithm.

At the $n^{th}$ iteration, the E-step generates the latent variable by estimating its posterior with the old model ($\theta^{n-1}$). In the original pseudo-labelling \cite{PseudoLabel}, pseudo-labels are generated according to its maximum predicted probability. For example in the binary setting, the pseudo-labels are generated using a fixed threshold value ($T$) which is normally set up as 0.5. Pseudo-labelling is equivalent to applying the plug-in principle \cite{EntropyMinimisation} to estimate the posterior of $Y'_U$ using its empirical estimation. Hence, the pseudo-labelling is the E-step:
\begin{equation}
    y'_u = \mathds{1} (\theta^{n-1}(x_u) > T=0.5)
    \label{equ:pseudo_label}
\end{equation}

At the M-step of iteration $n$, $Y'_U$ is fixed and we update the old model $\theta^{n-1}$ using the estimated latent variables (pseudo-labels $Y'_U$) and the data $X$ including both the labelled and unlabelled data:
\begin{equation}
    \theta^n := \argmax_\theta p(\theta^{n_1} | X, \theta^{n-1}, Y'_n)
    \label{equ:m_step_final}
\end{equation}

Eq.\ref{equ:m_step_final} can be solved by stochastic gradient descent given an objective function. We use Dice loss function ($f_{dice}(.)$) \cite{dice_loss} as the objective function and we weight the unsupervised learning with a hyper-parameter $\alpha$, the Eq.\ref{equ:m_step_final} and Eq.\ref{equ:pseudo_label} can thereby be extended to a loss function over the whole dataset (supervised learning part $L_L$ and unsupervised learning part $L_U$):

\begin{equation}
    \mathcal{L}_{SegPL} = \underbrace{\alpha \frac{1}{N-L} \sum_{u=L+1}^{N} f_{dice}(\theta^{n-1}(x_u), y'_u)}_{\mathcal{L}_U} + \underbrace{\frac{1}{L} \sum_{l=1}^{L} f_{dice}(\theta^{n-1}(x_l), y_l)}_{\mathcal{L}_L}
    \label{equ:pseudo_label_loss}
\end{equation}


\section{Generalisation of Pseudo Labels via Variational Inference for Segmentation}
We now derive a generalisation of SegPL under the same EM framework. As shown in the E-step in Eq.\ref{equ:pseudo_label}, the pseudo-labelling depends on the threshold value $T$, so finding the pseudo-label $y'_u$ is the same with finding the threshold value $T$. We therefore can recast the learning objective of EM as estimating the likelihood $P(X, T, \theta)$ w.r.t $\theta$. The benefit of flexible $T$ is that different stages of training might need different optimal $T$ to avoid wrong pseudo labels which cause noisy training\cite{defense_pseudo_label}. The E-step is now estimating the posterior of $T$. According to the Bayes' rule, the posterior of $T$ is $p(T | X, \theta) \propto p(X |\theta, T)p(T)$. The E-step at iteration $n$ is:
\begin{equation}
    p(T_n=i | X, \theta^{n-1}) = \frac{\prod_{u=L+1}^{N} p(x_u |\theta^{n-1}, T_n=i)p(T_n=i)}{\sum_{j \in [0, 1]} \prod_{u=L+1}^{N} p(x_u |\theta^{n-1}, T_n=j)p(T_n=j)}
    \label{equ:e_step_full}
\end{equation}
As shown in Eq.\ref{equ:e_step_full}, the denominator is computationally intractable as there are infinite possible values between 0 and 1. This also confirms that the empirical estimation in Eq.\ref{equ:pseudo_label} is actually necessary although not optimal. We use variational inference to approximate the true posterior of $p(T|X)$. We use new parameters $\phi$ \footnote{$\phi$: Average pooling layer followed by one 3 x 3 convolutional block including ReLU and normalisation layer, then two 1 x 1 convolutional layers for $\mu$ and $Log(\sigma)$} to parametrize $p(T|X)$ given the image features, assuming that $T$ is drawn from a Normal distribution between 0 and 1:
\begin{equation}
    (\mu, Log(\sigma)) = \phi(\theta(X))
    \label{equ:t_model}
\end{equation}
\vspace{-10pt}
\begin{equation}
    p(T|X, \theta, \phi) \approx  \mathcal{N}(\mu, \sigma)
    \label{equ:pt_parametrization}
\end{equation}
Following \cite{VAE}, we introduce another surrogate prior distribution over $T$ and denote this as $q(\beta), \beta \sim \mathcal{N}(\mu_\beta, \sigma_\beta)$. The E-step now minimises $KL(p(T) || q(\beta))$. The M-step is the same with Eq.\ref{equ:pseudo_label_loss} but with a learnt threshold $T$. The learning objective $P(X, T, \theta)$ has supervised learning $P(X_L, T, \theta)$ and unsupervised learning part $P(X_U, T, \theta)$. Although $P(X_L, T, \theta)$ is not changed, $P(X_U, T, \theta)$ now estimates the Evidence Lower Bound (ELBO):
\begin{equation}
    Log (P(X_U, T, \theta, \phi)) \geq  \sum^{N}_{u=L+1} \mathbb{E}_{T \sim P(T)} [Log (P(y_u | x_n, T, \theta, \phi))] - KL(p(T) || q(\beta))
    \label{equ:elbo}
\end{equation}
Where $Log$ is taken for expression simplicity in Eq.\ref{equ:elbo}, $y_n$ is the prediction of unlabelled image $x_u$. The final loss function is:
\begin{equation}
    \mathcal{L}^{VI}_{SegPL} = \mathcal{L}_{SegPL} + \underbrace{Log(\sigma_\beta) - Log(\sigma) + \frac{\sigma^2 + (\mu-\mu_\beta)^2}{2*(\sigma_\beta)^2} - 0.5}_{KL(p(T) || q(\beta)), \beta \sim \mathcal{N}(\mu_\beta,\,\sigma_\beta)}
    \label{equ:variational_loss}
\end{equation}
Where $\mathcal{L}_{SegPL}$ can be found in Eq.\ref{equ:pseudo_label_loss} but with a learnt threshold $T$. Different data sets need different priors. Priors are chosen empirically for different data sets. $\beta \sim \mathcal{N}(0.5, 0.1)$ for BRATS and $\beta \sim \mathcal{N}(0.4, 0.1)$ for CARVE. 

\section{Experiments \& Results}
\label{sec:exps}
\textbf{Segmentation Baselines \footnote{Training: Adam optimiser\cite{adam}. Our code is implemented using Pytorch 1.0\cite{pytorch}, trained with a TITAN V GPU. See Appendix for details of hyper-parameters.}} We compare SegPL to both supervised and semi-supervised learning methods. We use U-net~\cite{Unet} in SegPL as an example of segmentation network. Partly due to computational constraints, for 3D experiments we used a 3D U-net with 8 channels in the first encoder such that unlabelled data can be included in the same batch. For 2D experiments, we used a 2D U-net with 16 channels in the first encoder. The first baseline utilises supervised training on the backbone and is trained with labelled data denoted as ``Sup''. We compared SegPL with state-of-the-art consistency based methods: 1) ``cross pseudo supervision'' or CPS \cite{cpl_cvpr_2021}, which is considered the current state-of-the-art for semi-supervised segmentation; 2) another recent state-of-the-art model ``cross consistency training'' \cite{cct_cvpr2020}, denoted as ``CCT'', due to hardware restriction, our implementation shares most of the decoders apart from the last convolutional block; 3) a classic model called ``FixMatch'' (FM) \cite{fixmatch2020}. To adapt FixMatch for a segmentation task, we added Gaussian noise as weak augmentation and ``RandomAug'' \cite{random_augmentation} for strong augmentation; 4) ``self-loop \cite{self_loop_uncertainty}'', which solves a self-supervised jigsaw problem as pre-training and combines with pseudo-labelling.

\textbf{CARVE 2014} The classification of pulmonary arteries and veins (CARVE) dataset \cite{carve2014} comprises 10 fully annotated non-contrast low-dose thoracic CT scans. Each case has between 399 and 498 images, acquired at various spatial resolutions ranging from (282 x 426) to (302 x 474). We randomly select 1 case for labelled training, 2 cases for unlabelled training, 1 case for validation and the remaining 5 cases for testing. All image and label volumes were cropped to 176 $\times$ 176 $\times$ 3. To test the influence of the number of labelled training data, we prepared four sets of labelled training volumes with differing numbers of labelled volumes at: 2, 5, 10, 20. Normalisation was performed at case wise. Data curation resulted in 479 volumes for testing, which is equivalent to 1437 images.

\textbf{BRATS 2018} BRATS 2018 \cite{brats2015} comprises 210 high-grade glioma and 76 low-grade glioma MRI cases. Each case contains 155 slices. We focus on multi-class segmentation of sub-regions of tumours in high grade gliomas (HGG). All slices were centre-cropped to 176 x 176. We prepared three different sets of 2D slices for labelled training data: 50 slices from one case, 150 slices from one case and 300 slices from two cases. We use another 2 cases for unlabelled training data and 1 case for validation. 50 HGG cases were randomly sampled for testing. Case-wise normalisation was performed and all modalities were concatenated. A total of 3433 images were included for testing.


\begin{table}[t]
\caption{Our model vs Baselines on a binary vessel segmentation task on 3D CT images of the CARVE dataset. Metric is Intersection over Union (IoU ($\uparrow$) in \%) . All 3D networks have 8 channels in the first encoder. Avg performance of 5 training.}
  \label{tab:carve}
  \centering
  \resizebox{0.9\textwidth}{!}
  {
  \begin{tabular}{ c  | c  | c | c | c | c | c}
    \hline
    \bfseries Data & \bfseries Supervised & \multicolumn{5}{c}{\bfseries Semi-Supervised} \\
    \hline
    \bfseries Labelled & \bfseries 3D U-net &\bfseries FixMatch  & \bfseries CCT &\bfseries CPS &\bfseries SegPL &\bfseries SegPL+VI\\
    \bfseries Volumes & \cite{Unet}(2015) & \cite{fixmatch2020}(2020) & \cite{cct_cvpr2020}(2020) & \cite{cpl_cvpr_2021}(2021) & (Ours, 2022) & (Ours, 2022)\\
    \hline
    2 & 56.79$\pm$6.44 & 62.35$\pm$7.87 & 51.71$\pm$7.31 & 66.67$\pm$8.16 & \textbf{\textcolor{blue}{69.44$\pm$6.38}} & \textbf{\textcolor{red}{70.65$\pm$6.33}}\\
    \hline
    5 & 58.28$\pm$8.85 & 60.80$\pm$5.74 & 55.32$\pm$9.05 & 70.61$\pm$7.09 & \textbf{\textcolor{red}{76.52$\pm$9.20}} & \textbf{\textcolor{blue}{73.33$\pm$8.61}}\\
    \hline
    10 & 67.93$\pm$6.19 & 72.10$\pm$8.45 & 66.94$\pm$12.22 & 75.19$\pm$7.72 & \textbf{\textcolor{blue}{79.51$\pm$8.14}} &\textbf{\textcolor{red}{79.73$\pm$7.24}}\\
    \hline
    20 & 81.40$\pm$7.45 & 80.68$\pm$7.36 & 80.58$\pm$7.31 & 81.65$\pm$7.51 & \textbf{\textcolor{blue}{83.08$\pm$7.57}} &\textbf{\textcolor{red}{83.41$\pm$7.14}}\\
    \hline
    \multicolumn{7}{c}{\bfseries Computational need} \\
    \hline
    Train(s) & 1014 & 2674 & 4129 & 2730 & 1601 & 1715\\
    \hline 
    Flops & 6.22 & 12.44 & 8.3 & 12.44 & 6.22 & 6.23\\
    \hline
    Para(K) & 626.74 & 626.74 & 646.74 & 1253.48 & 626.74 & 630.0 \\
    \hline
    \end{tabular}
    }
\end{table}

\subsection{SegPL Outperforms Baselines with Less Computational Resource and Less Training Time}
\begin{figure}[t]
    \centering
    \begin{center}
        \includegraphics[width=0.75\textwidth]{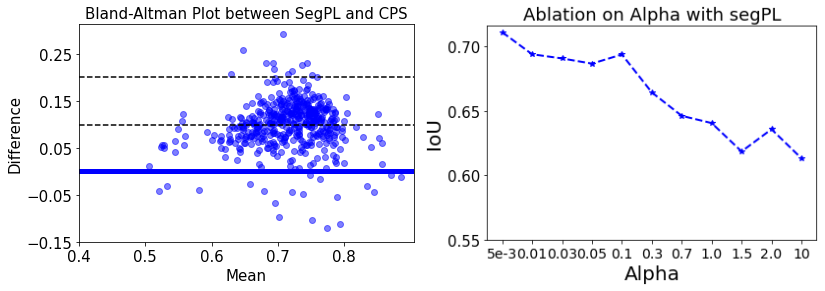}
    \end{center}
    \caption{Left: SegPL statistically outperforms the best performing baseline CPS when trained on 2 labelled volumes from the CARVE dataset. Each data point represents a single testing image. Right: Ablation study on the $\alpha$ which is the weight on the unsupervised learning part. Different data might need different $\alpha$.}
    \label{fig:iou_analysis}
\end{figure}

\begin{figure}[t]
    \centering
    \begin{center}
        \includegraphics[width=0.75\textwidth]{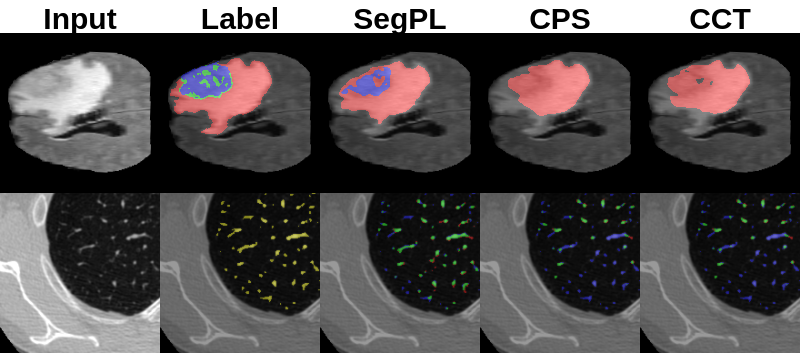}
    \end{center}
    \caption{Row1: BRATS with 300 labelled slices. Red: whole tumour. Green: tumour core. Blue: enhancing tumour core. Row2: CARVE with 5 labelled volumes. Red: false positive. Green: true positive. Blue: false negative. Yellow: ground truth.}
    \label{fig:visual_result}
\end{figure}

As shown in Tab.\ref{tab:brats}, Tab.\ref{tab:carve} and Fig.\ref{fig:visual_result}, SegPL consistently outperformed the state-of-the-art semi-supervised segmentation methods and the supervised counterpart for different labelled data regimes, on both 2D multi-class and 3D binary segmentation tasks. The Bland-Altman plot in Fig.\ref{fig:iou_analysis}.Left shows SegPL performs significantly better than the best performed baseline CPS in an extremely low data regime (2 labelled volume on CARVE). We also applied Mann Whitney test on results when trained with 2 labelled volume and CARVE and the p-value was less than 1e-4. As shown in bottom three rows in Tab.\ref{tab:carve}, SegPL also is the most close one to the supervised learning counterpart in terms of computational cost, among all of the tested methods. As shown in Tab.\ref{tab:brats}\footnote{\textcolor{blue}{blue: 2nd best}. \textcolor{red}{red: best}}, SegPL-VI further improves segmentation performance on the MRI based brain tumour segmentation task. It also appears that the variational inference is sensitive to the data because we didn't observe obvious performance gain with SegPL-VI on CARVE. The performance of SegPL-VI on CARVE could be further improved with more hyper-parameter searching.


\begin{table}[t]
\caption{Our model vs Baselines on multi-class tumour segmentation on 2D MRI images of BRATS. Metric is Intersection over Union (IoU ($\uparrow$) in \%). All 2D networks have 16 channels in the first encoder. Avg performance of 5 runs.}
  \label{tab:brats}
  \centering
  \resizebox{0.9\textwidth}{!}
  {
  \begin{tabular}{ c  | c  |  c | c | c | c | c}
    \hline
    \bfseries Data & \bfseries Supervised & \multicolumn{5}{c}{\bfseries Semi-Supervised} \\
    \hline
    \bfseries Labelled & \bfseries 2D U-net &\bfseries Self-Loop &\bfseries FixMatch &\bfseries CPS &\bfseries SegPL &\bfseries SegPL+VI\\
    \bfseries Slices & \cite{Unet}(2015) & \cite{self_loop_uncertainty}(2020) & \cite{fixmatch2020}(2020) & \cite{cpl_cvpr_2021}(2021) & (Ours, 2022) & (Ours, 2022)\\
    \hline
    50 & 54.08$\pm$10.65 & 65.91$\pm$10.17 & 67.35$\pm$9.68 & 63.89$\pm$11.54 & \textcolor{blue}{\textbf{70.60$\pm$12.57}} &\textcolor{red}{\textbf{71.20$\pm$12.77}}\\
    \hline
    150 & 64.24$\pm$8.31 & 68.45$\pm$11.82 & 69.54$\pm$12.89 & 69.69$\pm$6.22 & \textcolor{blue}{\textbf{71.35$\pm$9.38}} & \textcolor{red}{\textbf{72.93$\pm$12.97}}\\
    \hline
    300 & 67.49$\pm$11.40 & 70.80$\pm$11.97 & 70.84$\pm$9.37 & 71.24$\pm$10.80 & \textcolor{blue}{\textbf{72.60$\pm$10.78}} & \textcolor{red}{\textbf{75.12$\pm$13.31}}\\
    \hline
    \end{tabular}
    }
\end{table}

\subsection{Ablation Studies} 
Ablation studies are performed using BRATS with 150 labelled slices. The suitable range for $\alpha$ is $5e-3$ to $0.1$ as shown in Fig.\ref{fig:iou_analysis}.Right. The other ablation studies can be found in the Appendix. We found that SegPL needs large learning rate from at least 0.01. We noticed that SegPL is not sensitive to the warm-up schedule of $\alpha$. The appropriate range of the ratio between unlabelled images to labelled images in each batch is between 2 to 10. Different data sets might need different ranges of $\alpha$ for optimal performances.

\begin{figure}[!tb]
   \begin{minipage}{0.48\textwidth}
     \centering
     \includegraphics[width=\linewidth]{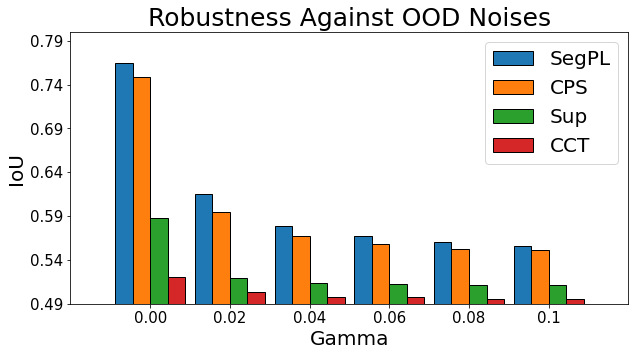}
         \caption{Robustness against out-of-distribution noise. Gamma is the strength of the out-of-distribution noises. Using 2 labelled volumes from CARVE.}\label{fig:out_of_distribution}
   \end{minipage}\hfill
   \begin{minipage}{0.48\textwidth}
     \centering
     \includegraphics[width=\linewidth]{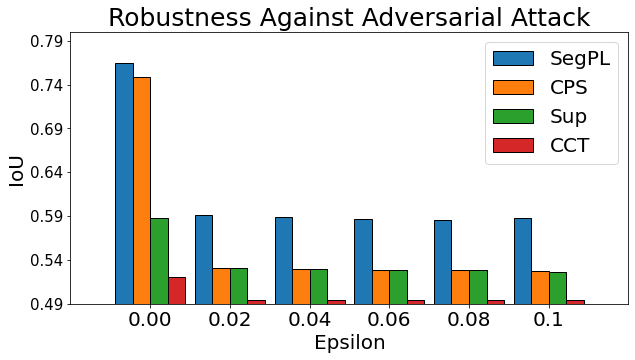}
     \caption{Robustness against adversarial attack. Epsilon is the strength of the FGSM\cite{adversarial_attack} attack. Using 2 labelled volumes from CARVE.}\label{fig:adversarial_attack}
   \end{minipage}
\end{figure}

 \subsection{Better Generalisation on Out-Of-Distribution (OOD) Samples and Better Robustness against Adversarial Attack} 
 We mimic OOD image noise resulting from variations in scan acquisition parameters and patient characteristics that are characteristic of medical imaging. We simulated OOD samples with unseen random contrast and Gaussian noise on CARVE and mix-up \cite{mixup} to create new testing samples. Specifically, for a given original testing image $x_t$, we applied random contrast and noise augmentation on $x_t$ to derive OOD samples $x'_t$. We arrived at the testing sample ($\hat{x_t}$) via $\gamma x'_t + (1 - \gamma) x_t$. As shown in Fig.\ref{fig:out_of_distribution}, as testing difficulty increases, the performances across all baselines drop exponentially. SegPL outperformed all of the baselines across all of the tested experimental settings. The findings suggest that SegPL is more robust when testing on OOD samples and achieves better generalisation performance against that from the baselines. As robustness has become important in privacy-preserving collaborative learning among hospitals, we also examined SegPL's robustness against adversarial attack using fast gradient sign method (FGSM)\cite{adversarial_attack}. With increasing strength of adversarial attack (Epsilon), all the networks suffered performance drop. Yet, as shown in Fig.\ref{fig:adversarial_attack}, SegPL suffered much less than the baselines.


\subsection{Uncertainty Estimation with SegPL-VI} 
SegPL-VI is trained with stochastic threshold for unlabelled data therefore not suffering from posterior collapse. Consequently, SegPL can generate plausible segmentation during inference using stochastic thresholds. We tested the uncertainty estimation of SegPL-VI with the common metric Brier score on models trained with 5 labelled volumes of CARVE dataset. We compared SegPL-VI with Deep Ensemble, which is regarded as a gold-standard baseline for uncertainty estimation \cite{deep_ensemble}\cite{evaluation_uncertainty}. Both SegPL-VI and Deep Ensemble achieved the same Brier score at 0.97, using 5 Monte Carlo samples.

\section{Related Work}
Although we are the first to solely use the original pseudo labelling in semi-supervised segmentation of medical images, a few prior works explored the use of pseudo labels in other forms. Bai \cite{ssl_cardiac} post-processed pseudo-labels via conditional random field. Wu \cite{SSS_MCT} proposed sharpened cycled pseudo labels along with a two-headed network. Wang \cite{SSS_CT_Attention} post-processed pseudo-labels with uncertainty-aware refinements for their attention based models.

\section{Conclusion}
We verified that the original pseudo-labelling \cite{PseudoLabel} is a competitive and robust baseline in semi-supervised medical image segmentation. We further interpret pseudo-labelling with EM and explore the potential improvement with variational inference following generalised EM. The most prominent future work is to improve the quality of the pseudo labels by exploring different priors (e.g. Beta, Categorical) or auto-search of hyper parameters. Other future works include investigating the convergence property of SegPL-VI and improving SegPL-VI at uncertainty quantification. Future studies can also investigate from the perspective of model calibration \cite{Xu_MIDL} to understand why pseudo labels work. SegPL and SegPL-VI can also be used in other applications such as classification or registration. 
%

\section{Acknowledgements}
We thank the anonymous reviewer 3 for the suggestion of using Beta prior and an interesting discussion about posterior collapse issue in uncertainty quantification. MCX is supported by GlaxoSmithKline (BIDS3000034123) and UCL Engineering Dean’s Prize. NPO is supported by a UKRI Future Leaders Fellowship (MR/S03546X/1). DCA is supported by UK EPSRC grants M020533, R006032, R014019, V034537, Wellcome Trust UNS113739. JJ is supported by Wellcome Trust Clinical Research Career Development Fellowship
209,553/Z/17/Z. NPO, DCA, and JJ are supported by the NIHR UCLH Biomedical Research Centre, UK.

\bibliographystyle{splncs04}
\bibliography{paper2505}
\end{document}


\title{Extended Appendix for Bayesian Pseudo Labels}

\titlerunning{Bayesian Pseudo Labels}
%
\author{
Mou-Cheng Xu\inst{1} \and
Yu-Kun Zhou\inst{1} \and
Chen Jin\inst{1}\and
Marius de Groot\inst{2} \and
Daniel C. Alexander\inst{1} \and
Neil P. Oxtoby\thanks{Joint Senior Authorships.} \inst{1} \and
Yipeng Hu \samethanks \inst{1} \and
Joseph Jacob \samethanks \inst{1}
}

%
\authorrunning{Xu, Zhou, Jin, de Groot, Alexander, Oxtoby, Hu, Jacob}
%


\institute{
Centre for Medical Image Computing, University College London, UK \and
GSK R \& D, Stevenage, UK 
}

\maketitle 

\section{Training Hyper-Parameters}
\begin{table}[!htb]
\caption{Hyper-parameters used across experiments. Different data might need different $\alpha$. 3D models need longer time to converge.}
    \centering
    \begin{tabular}{c c c c c c }
        \hline
        Data & Batch Size & Learning rate & Steps & $\alpha$ & Unlabelled/labelled\\
        \hline
        BRATS  & 2 & 0.03 & 200 & 0.05 & 5\\
        \hline
        CARVE & 2 & 0.01 & 800 & 1.0 & 4\\
        \hline
    \end{tabular}
    \label{tab:hyperparameters}
\end{table}

\section{Results of Ablation Studies}
\begin{figure}[!ht]
    \centering
    \begin{center}
        \includegraphics[width=\textwidth]{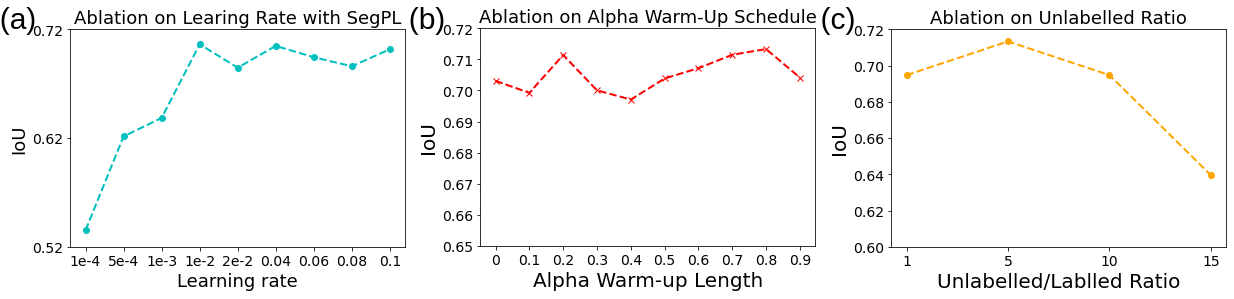}
    \end{center}
    \caption{Ablation studies are performed on hyper-parameters trained on BRATS with 150 labelled slices. a) Ablation on learning rate. Learning rate should be at least 0.01. We found that other baselines also needed large learning rate. b) Ablation on warm-up schedule of $\alpha$ from 0 to final $\alpha$ value. x axis is the length of linear warming-up of $\alpha$ in terms of whole steps. It appears that SegPL is not sensitive to warm-up schedule of $\alpha$. c) Ablation on the ratio between unlabelled images to labelled images in each batch. The suitable range of unlabelled/labelled ratio is quite wide and between 1 to 10.}
    \label{fig:ablation_study_more}
\end{figure}

\section{Brier Score}
\begin{equation}
\label{brier}
Brier = \frac{1}{HW} \sum_{i=1}^{H} \sum_{j=1}^{W} (p_{ij} - y_{ij})^2
\end{equation} 
Where, $y_{ij}$ is the ground truth label at pixel at location i, j, $y_{ij}$ is 1 for foreground pixel and $y_{ij}$ is 0 for background pixel. $p_{ij}$ is the predicted probability of the pixel being the foreground pixel.

\section{Differences between SegPL and Other methods}
\begin{figure}[ht!]
    \centering
        \includegraphics[width=\textwidth]{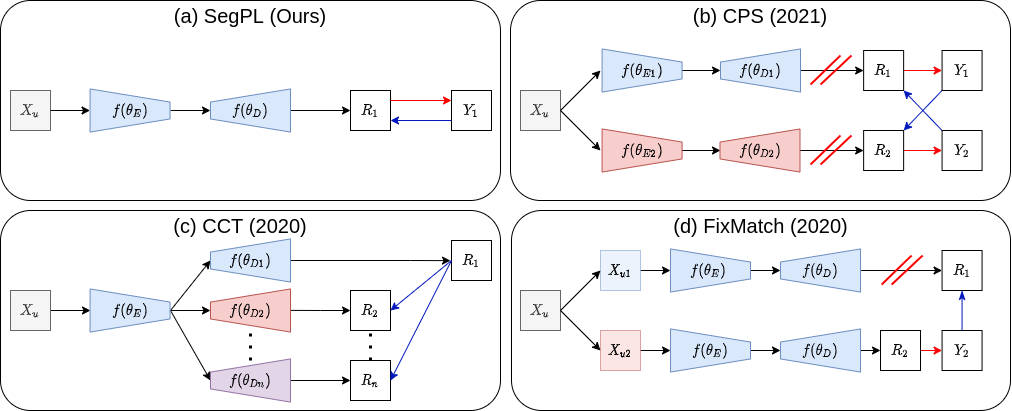}
    \caption{Comparisons between SegPL and popular self-ensemble semi-supervised segmentation methods on unlabelled images. (a) Proposed SegPL. (b) CPS (cross-pseudo-supervision) (c) CCT (cross-consistency training). (d) FixMatch uses different augmentation at the input level. $\rightarrow$: data flow. \textcolor{red}{$\rightarrow$}: generate pseudo label. \textcolor{blue}{$\rightarrow$}: supervision with loss function. \textcolor{red}{//}: stopping gradients.}
    \label{fig:method}
\end{figure}

\textbf{SegPL vs CPS } CPS uses two models and cross-pseudo-supervision, as illustrated in Fig\ref{fig:method}, in which the first pseudo label is from the second model and the second pseudo label is from the first model. SegPL uses a single model that requires a single forward pass. 

\textbf{SegPL vs FixMatch } The main difference between SegPL and FixMatch relates to the augmentation of the input images. In FixMatch (see Fig.\ref{fig:method}), the unlabelled image is randomly augmented twice, once weakly and once strongly. Each augmented image requires one forward pass and the pseudo label of FixMatch is generated from the weakly augmented image. SegPL however does not use data augmentation.
$Pseudo \ Label = \mathds{1} (Model(x) > T=0.5)$

\section{Pre processing Of the Labels For Multi-Class Segmentation}
\begin{figure}[ht]
\centering
\includegraphics[width=\textwidth]{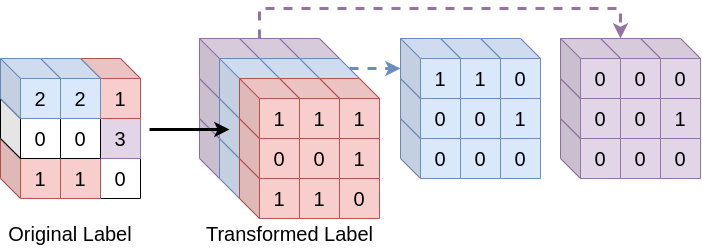}
\caption{An example of the pre-processing of one label of BRATS. 3: enhancing tumour core. 2: tumour core containing enhancing tumour core. 1: whole tumour containing class 2 and 3. 0: healthy tissues. Different colours represent different classes.}
\label{fig:multi_to_binary_example}
\end{figure}

\begin{figure}[ht]
\centering
\includegraphics[width=\textwidth]{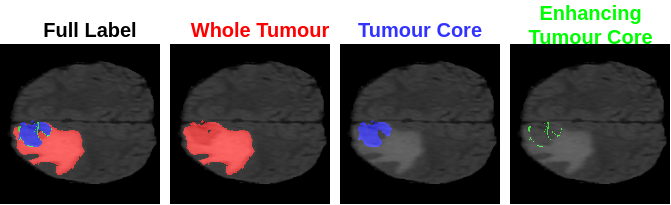}
\caption{Label fusion and binarized labels. Red: whole tumour including tumour core. Blue: tumour core including enhancing tumour core. Green: enhancing tumour core. Segmentation of each tumour class is a binary segmentation.}
\label{fig:brats_label}
\end{figure}

The pre-processing of the labels of BRATS has two steps: 1) label fusion to mitigate the severe class imbalance between the minority tumour classes and the majority background healthy tissue class; 2) turning a multi-class label into multiple binary labels, for each binary prediction, we can use Sigmoid followed by confidence thresholding for pseudo labelling. Here we show two examples of label pre-processing, one is an abstract example (Fig.\ref{fig:multi_to_binary_example}) and the other one is a real example (Fig.\ref{fig:brats_label}).

\section{Segmentation Performance and Sizes of Foreground Areas}
\begin{figure}[!hb]
\centering
\includegraphics[width=\textwidth]{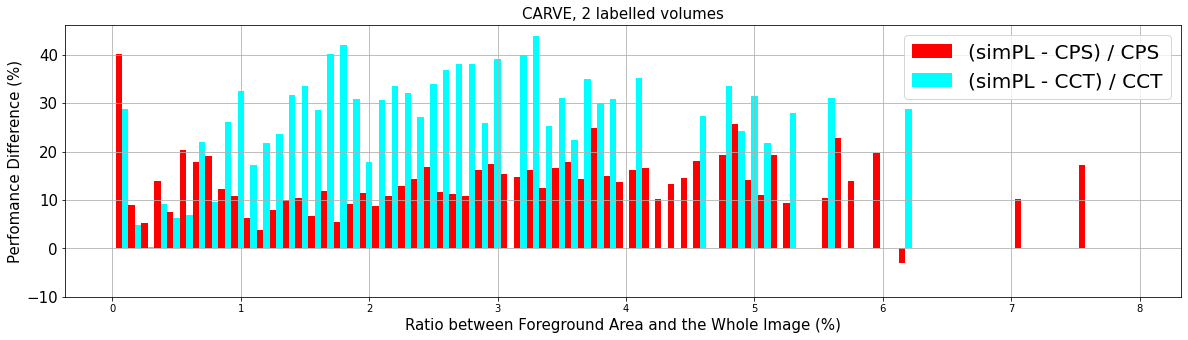}
\caption{How SegPL outperforms baselines at each size of foreground area. When the foreground area is extremely small (very left bin, less than 0.1$\%$), SegPL outperforms CPS by 40$\%$. It is also easy to see that, segmentation of vessels is a very hard task in general, the regions of the interest is less than 8$\%$ of the whole images. The segmentation of vessels suffers from severe class imbalance between the foreground and the background.}
\label{fig:iou_foregorund_size}
\end{figure}

\section{Temperature Scaling}
\begin{figure}[ht]
    \centering
        \includegraphics[width=0.9\textwidth]{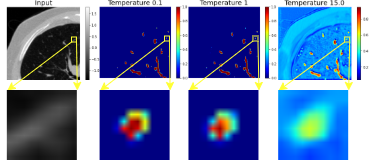}
    \caption{Column 1: input. Column 2: temperature 0.1. Column 3: temperature 1.0. Column 4: temperature 15.0. The colour bar has deep red as most confident (1) and deep blue as least confident (0).}
    \label{fig:conf_visual}
\end{figure}

\section{More Visual Results}
\begin{figure}[!hb]
    \centering
    \begin{center}
        \includegraphics[width=\textwidth]{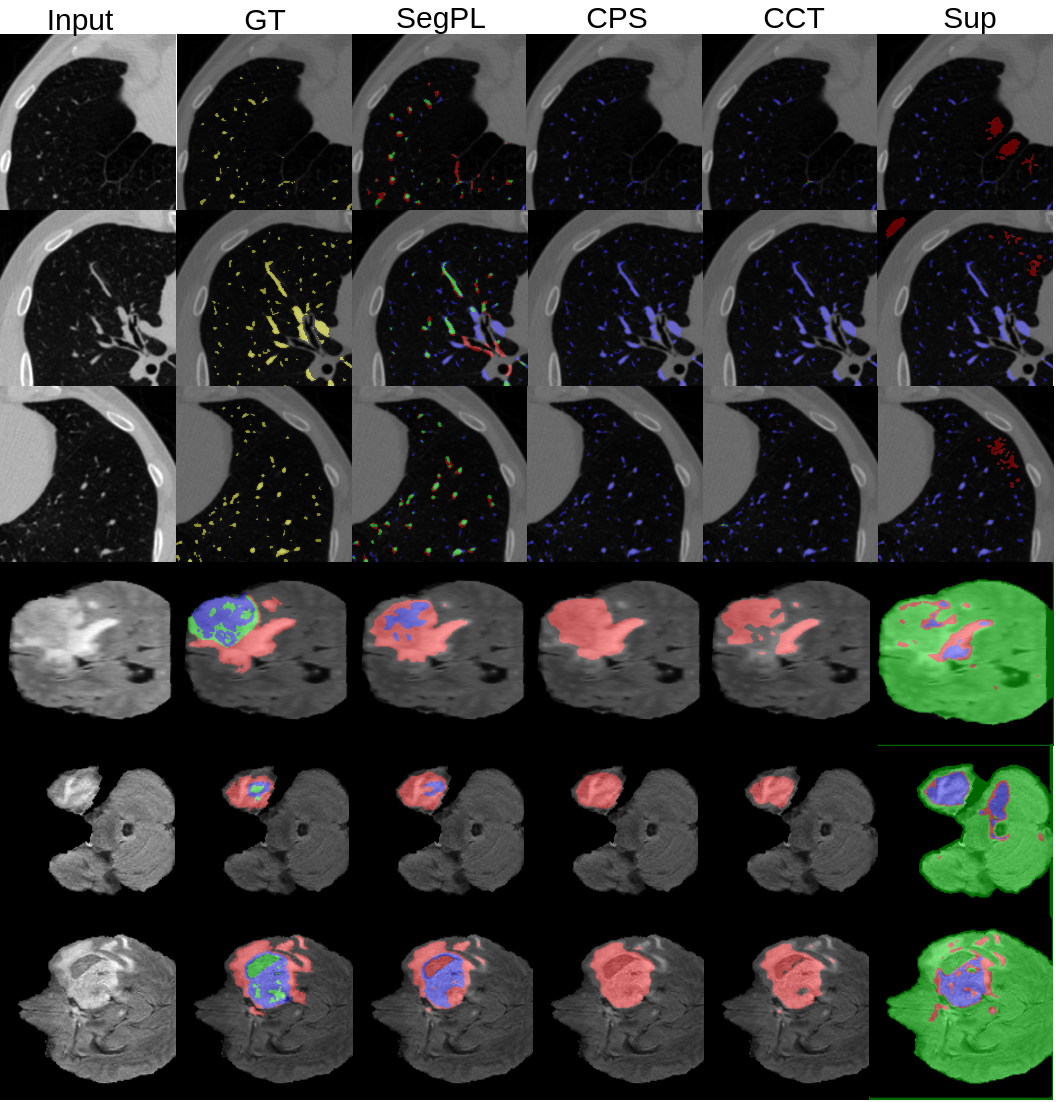}
    \end{center}
    \caption{Row1-3: CARVE trained with 5 labelled volumes. Red: false positive. Green: true positive. Blue: false negative. Yellow: ground truth. Row4-6: BRATS trained with 300 labelled slices. Red: whole tumour. Green: tumour core. Blue: enhancing tumour core. GT: Ground truth. CPS: cross pseudo labels (CVPR 2021). CCT: cross consistency training (CVPR 2020). Sup: supervised training.}
    \label{fig:visual_result_more}
\end{figure}

\section{Results Analysis}
\begin{figure}[ht]
\centering
\includegraphics[width=0.7\textwidth]{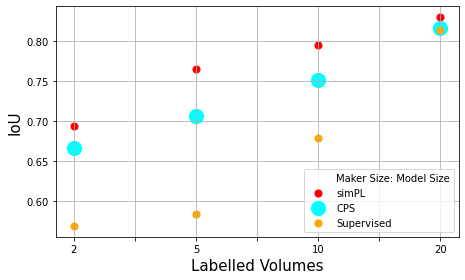}
\caption{Main results on the model size and the performances among different methods on CARVE data set. segPL outperforms baselines while uses less computational power.}
\label{fig:main_result}
\end{figure}

\begin{figure}[ht]
    \centering
        \includegraphics[width=0.8\textwidth]{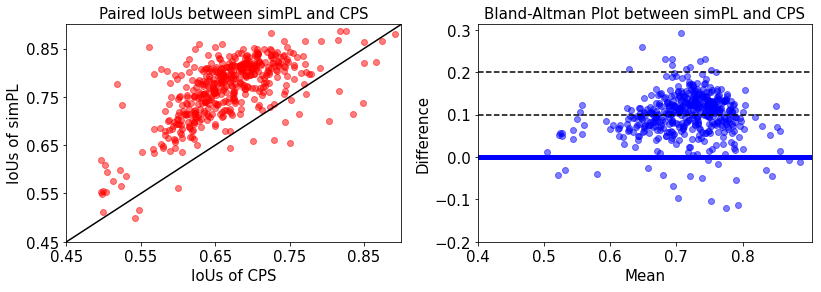}
    \caption{Left: SegPL statistically outperforms the best performing baseline CPS when trained on 2 labelled volumes from the CARVE dataset. Each data point represents a single testing image. Right: Ablation study on the $\alpha$ which is the weight on the unsupervised learning part. Different data might need different $\alpha$.}
    \label{fig:iou_analysis}
\end{figure}